\documentclass[10pt,conference]{IEEEtran}
\IEEEoverridecommandlockouts
\usepackage{cite}
\usepackage{amsmath,amssymb,amsfonts}
\usepackage{algorithmic}
\usepackage{graphicx}
\usepackage{textcomp}
\usepackage{xcolor}
\usepackage{algorithm}
\usepackage{array}
\usepackage[caption=false,font=normalsize,labelfont=sf,textfont=sf]{subfig}

\usepackage{url}
\usepackage{xurl}
\usepackage{hyperref}

\usepackage{cleveref}
\usepackage{xcolor}
\usepackage{tikz}
\usepackage{pifont}

\hypersetup{colorlinks=true, linkcolor=blue, anchorcolor=red, citecolor=blue, filecolor =red, pdfauthor=author}
\usepackage{xcolor}
\usepackage{color, colortbl}
\definecolor{rosso}{RGB}{249,128,115}

\usepackage{utfsym}
\usepackage{pifont} % puoi tenerlo se usato altrove
\DeclareRobustCommand{\cmark}{\usym{1F5F8}}
\DeclareRobustCommand{\xmark}{\scalebox{0.95}{\usym{2613}}}
\usepackage{pgfplots}
\pgfplotsset{compat=1.18}
\usepackage{layouts}
\usepackage{comment}
\usepackage{threeparttable}
\usepackage{bm}
\usepackage{enumitem}
\usepackage{setspace}
\usepackage{times}
\usepackage{balance}  
\usepackage{soul}               
\usepackage{url}
\usepackage{textcomp}
\usepackage{mathrsfs}
\usepackage{dcolumn}
\usepackage{color}
\usepackage{array}
\usepackage{textcomp}
\usepackage{siunitx}
\usepackage{lipsum}
\definecolor{Gray}{gray}{0.9}

\definecolor{lightblue}{HTML}{E6E5FD}

\usepackage{longtable}
\usepackage{pdflscape}
\usepackage{lineno}
\usepackage{booktabs}
\usepackage{tabularx}
\usepackage{placeins}
\usepackage[table]{xcolor}

\definecolor{pv}{HTML}{4F6272}       % blu per PV
\definecolor{weath}{HTML}{C25E5E}     % rosso per Weather
\definecolor{fused}{HTML}{E3B04B}  % arancione per Fused
\definecolor{forecast}{HTML}{5C8D4C}% verde per Weather Forecast

\definecolor{lightblue}{HTML}{E6E5FD}

\usepackage{comment}
\usepackage{url}

\def\BibTeX{{\rm B\kern-.05em{\sc i\kern-.025em b}\kern-.08em
    T\kern-.1667em\lower.7ex\hbox{E}\kern-.125emX}}

\DeclareRobustCommand{\STAT}{\textcolor{blue}{$\blacksquare$}}
\DeclareRobustCommand{\REC}{\textcolor{teal}{$\blacktriangle$}}
\DeclareRobustCommand{\CONV}{\textcolor{orange}{$\bullet$}}
\DeclareRobustCommand{\TRANS}{\textcolor{purple}{$\blacklozenge$}}
\DeclareRobustCommand{\LLMcat}{\textcolor{red}{$\star$}}

\begin{document}
\bstctlcite{BSTcontrol}

\title{Probabilistic NDVI Forecasting from Sparse Satellite Time Series and Weather Covariates
\thanks{Corresponding author: Matteo Tortora, E-mail: matteo.tortora@unige.it, Address: Via all’Opera Pia 11a, 16145 Genoa, Italy.}
}

\author{
\IEEEauthorblockN{
Irene Iele\IEEEauthorrefmark{1}\IEEEauthorrefmark{4},
Giulia Romoli\IEEEauthorrefmark{2}\IEEEauthorrefmark{4},
Daniele Molino\IEEEauthorrefmark{1},
Elena {Mulero Ayllón}\IEEEauthorrefmark{1},
Filippo Ruffini\IEEEauthorrefmark{1}\IEEEauthorrefmark{2}, \\
Paolo Soda\IEEEauthorrefmark{1}\IEEEauthorrefmark{2},
Matteo Tortora\IEEEauthorrefmark{3}
}
\IEEEauthorblockA{\IEEEauthorrefmark{1}
Unit of Artificial Intelligence and Computer Systems, Università Campus Bio-Medico di Roma, Italy \\
%Email: \{irene.iele, daniele.molino, e.muleroayllon, p.soda\}@unicampus.it
}
\IEEEauthorblockA{\IEEEauthorrefmark{2}
Department of Diagnostics and Intervention, Biomedical Engineering and Radiation Physics, Umeå University, Sweden
\\ 
%Email: \{giulia.romoli, filippo.ruffini, paolo.soda\}@umu.se
}

\IEEEauthorblockA{\IEEEauthorrefmark{3}
Department of Naval, Electrical, Electronics and Telecommunications Engineering, University of Genoa, Italy\\
Email: matteo.tortora@unige.it
}

\IEEEauthorblockA{\IEEEauthorrefmark{4}These authors contributed equally.}
}

\maketitle

\begin{abstract}
Short-term forecasting of vegetation dynamics is a key enabler for data-driven decision support in precision agriculture.
Normalized Difference Vegetation Index (NDVI) forecasting from satellite observations, however, remains challenging due to sparse and irregular sampling caused by cloud masking, as well as the heterogeneous climatic conditions under which crops evolve.
In this work, we propose a probabilistic forecasting framework for field-level NDVI prediction under sparse, irregular clear-sky acquisitions.
The architecture separates the encoding of historical NDVI and meteorological observations from future exogenous covariates, fusing both representations for multi-step quantile prediction.
To address irregular revisit patterns and horizon-dependent uncertainty, we introduce a temporal-distance weighted quantile loss that aligns the training objective with the effective forecasting horizon.
In addition, we incorporate cumulative and extreme-weather feature engineering to capture delayed meteorological effects relevant to vegetation response.
Experiments on European satellite data show that the proposed approach outperforms statistical, deep learning, and time-series baselines on both pointwise and probabilistic evaluation metrics.
Ablation studies confirm that target history is the primary driver of performance, with meteorological covariates providing additional gains in the full multimodal setting.
The code is available at \url{https://github.com/arco-group/ndvi-forecasting}.
\end{abstract}

\begin{IEEEkeywords}
NDVI Forecasting, Multimodal Fusion, Irregular Sampling, Quantile Forecasting, Remote Sensing, Weather Covariates
\end{IEEEkeywords}

\section{Introduction}
Precision agriculture aims to improve the efficiency and sustainability of farming practices by enabling timely, data-driven interventions tailored to local conditions. 
Short-term forecasts are particularly relevant for decisions such as irrigation scheduling, fertilization, and stress mitigation, where reducing uncertainty can support proactive crop management. 
In this context, Artificial Intelligence (AI) has become a key enabling technology across the agricultural production chain~\cite{adewusi2024,upadhyay2024,gangwani2024,tortora2025towards}, including recent agentic formulations for automatic farming systems \cite{bonifacio2026agentic}, and has shown increasing relevance in several decision-support settings involving heterogeneous or temporally evolving data~\cite{cordelli2020time,tortora2021deep,tortora2022pytrack,furia2023exploring,liu2021exploring,nibid2023deep,coser2025deep,cordelli2024machine,ayllon2025can}, supporting the shift from uniform to adaptive management strategies~\cite{misra2024}.
These challenges motivate models that account for irregular sampling and exploit complementary information sources, such as meteorological covariates. 
Similar principles have been explored in multimodal learning settings~\cite{iele2026hybrid,tortora2023matnet,ayllon2026context,tortora2023radiopathomics}.

Satellite remote sensing provides scalable and non-invasive monitoring of crop dynamics over large areas, complementing in-field sensors that are often impractical to deploy at scale~\cite{zhang2024,pal2025}. 
Multispectral observations enable the computation of Vegetation Indices (VIs), which describe vegetation properties related to phenology and stress response~\cite{xue2017,gong2024}. 
Among them, the Normalized Difference Vegetation Index (NDVI) is widely used as a proxy for vegetation greenness and canopy development. 
However, operational NDVI forecasting remains challenging: clear-sky Sentinel-2 observations are sparse and irregular due to revisit schedules and cloud masking, and forecasting performance degrades across heterogeneous agro-climatic conditions~\cite{sishodia2020,demissie2026}. 
These challenges motivate models that account for irregular sampling and exploit complementary information sources, such as meteorological covariates.
VI forecasting can be addressed at different output granularities. 
Pixel-level approaches forecast dense vegetation maps from spatio-temporal image inputs, as in ContextFormer~\cite{benson2024} and VegeDiff~\cite{zhao2025}; however, they require dense spatial inputs and are evaluated under image forecasting criteria. 
In contrast, this work focuses on field-level NDVI forecasting, where vegetation indices are aggregated over parcels or regions and modeled as sparse trajectories, a setting suited to short-term operational decision support. 
Existing field-level methods include recurrent models with meteorological covariates~\cite{ahmad2023,cavalli2021}, graph-based extensions~\cite{beyer2023}, alignment-based approaches~\cite{zhao2021}, and recent studies on robustness across European climates~\cite{farbo2024}. 
Nevertheless, many approaches rely on regular composites or implicitly densify the signal, and robustness under sparse and irregular clear-sky acquisitions remains underexplored.
This work introduces a probabilistic forecasting framework for field-level NDVI prediction across European ecozones and growing seasons. 
We employ a self-attention transformer that integrates historical NDVI with historical and future weather covariates to predict NDVI quantiles up to 14 days ahead, explicitly addressing sparse and irregular clear-sky sampling.

The main contributions are summarized as follows: 
\begin{itemize}
\item We propose a transformer-based quantile model for field-level NDVI forecasting under sparse and irregular clear-sky observations, jointly exploiting historical NDVI, historical weather, and future weather covariates; 
\item We introduce a temporal-distance weighted quantile loss to handle variable forecasting horizons induced by irregular revisit patterns; 
\item  We incorporate cumulative and extreme-weather features to capture delayed meteorological effects on vegetation response;
\item We validate the approach across European ecozones and growing seasons against statistical, deep learning, and recent time-series baselines, with ablation studies quantifying the contribution of each component.
\end{itemize}
\begin{figure*}
    \centering
\includegraphics[width=0.95\textwidth]{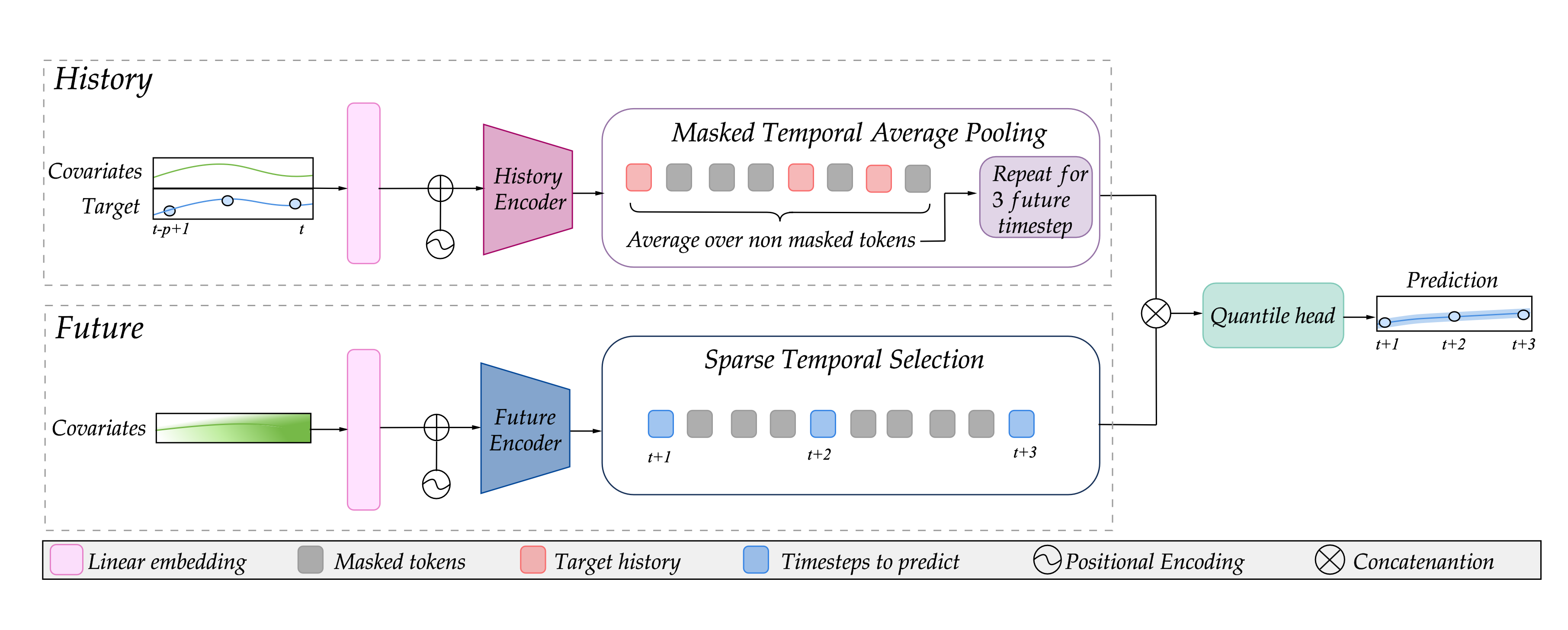}
    \caption{Transformer-based probabilistic forecasting architecture with decoupled history and future branches. Historical inputs are encoded and aggregated via temporal pooling, while future covariates are encoded and sparsely selected at acquisition times. 
    The fused representation predicts multi-step NDVI quantiles at levels $q \in \{0.1, 0.5, 0.9\}$.}
 \label{fig:methods}
\end{figure*}

\section{Materials}
\label{sec:materials}
In this study, we use the GreenEarthNet dataset~\cite{benson2024}, which provides spatio-temporal data cubes coupling cloud-masked Sentinel-2 image sequences with meteorological time series.
The dataset includes 24{,}061 data cubes over Europe from 2017 to 2022; each cube comprises 30 cloud-masked Sentinel-2 images at 5-day intervals ($128 \times 128$ pixels spanning $2.56 \times 2.56$ km) and 150 daily meteorological observations.
Following the official split, the train subset (2017--2019) is used for model training and the validation subset (2020) as the test set for final evaluation.

The Sentinel-2 images include blue (B02), green (B03), red (B04), and near-infrared (B8A) bands at 20\,m resolution, from which NDVI is computed as:
\begin{equation}
     \text{NDVI} = \frac{\text{B8A} - \text{B04}}{\text{B8A} + \text{B04}}
     \label{eq:NDVI}
\end{equation}
A pre-processing pipeline extracts NDVI time series from the data cube and enriches them with meteorological covariates.
For each cube, cloudy pixels are excluded via the dataset cloud mask, and NDVI is computed by averaging the remaining valid pixels at each acquisition timestamp.
When a Sentinel-2 overpass is fully obscured by clouds, no valid NDVI measurement is available and the acquisition is treated as missing.
This approach preserves the physical reliability of the vegetation signal at the cost of an irregular temporal sampling.

Formally, we consider a supervised time-series forecasting problem.
Let $\{(y_t, \tau_t, x_t)\}_{t=1}^{T}$ denote the dataset, where $y_t\in\mathbb{R}$ is the clear-sky NDVI target at the $t$-th observation, $\tau_t$ is its acquisition time, and $x_t\in\mathbb{R}^{d}$ collects the corresponding exogenous covariates.
Given a reference index $t$, the model receives the target history  $\mathbf{y}_{t-p+1:t}=(y_{t-p+1},\dots,y_t)$, the historical  covariates $\mathbf{x}_{t-p+1:t}=(x_{t-p+1},\dots,x_t)$, and the  future exogenous covariates $\mathbf{x}_{t+1:t+h}=(x_{t+1},\dots, x_{t+h})$, and predicts the future target sequence $\hat{\mathbf{y}}_{t+1:t+h}=(\hat{y}_{t+1}, \dots,\hat{y}_{t+h})$.
Here, $p$ denotes the number of historical clear-sky observations, and $h$ denotes the number of future clear-sky acquisitions to be predicted. 
Since indices refer to ordered clear-sky observations rather than fixed temporal steps, the spacing
\(\tau_{t+1} - \tau_t\) between consecutive targets is generally irregular, and the effective forecasting horizon
\(\tau_{t+h} - \tau_t\) varies across samples.
Forecasting samples are generated via a sliding-window strategy defined over the sequence of available NDVI observations, which occur only at Sentinel-2 overpass timestamps.
No target values exist between consecutive acquisitions, so the resulting series is sparse and irregularly sampled.
The history and forecasting horizons are set to \(p=3\) and \(h=3\), with a shift of four observations, corresponding to approximately 14 days depending on the satellite revisit schedule.
This horizon is appropriate for operational decision support, covering irrigation scheduling, fertilization, and crop management interventions before uncertainty accumulates excessively~\cite{hatfield2015temperature, cavalli2021}.

Temporal information is encoded via Fourier-based cyclical representations of the day of the year.
Sine and cosine components are computed up to the third harmonic, capturing sub-seasonal patterns without introducing high-frequency components that could amplify noise.

Meteorological variables are incorporated as exogenous covariates and enriched through cumulative feature engineering to capture delayed and extreme-weather effects on vegetation dynamics.
Nine derived features are computed from precipitation and temperature using two complementary aggregation strategies. \\
\emph{a) Between-target:}
Three features aggregate daily meteorological variables over the variable-length interval between two consecutive target observations: cumulative rainfall, the number of cold days ($T < 10^\circ$C), and the number of hot days ($T > 30^\circ$C).
% Since these features are computed by aggregating meteorological variables over the temporal intervals delimited by consecutive target observations, their values depend on both the duration and the conditions within each interval, and therefore vary across samples.

\emph{b) Rolling-window:}
Six features are computed using rolling aggregation over fixed temporal windows to capture short- and medium-term weather effects independently of sampling irregularity: cumulative rainfall and cold ($T < 10^\circ$C) and hot ($T > 30^\circ$C) days count over 7- and 14-day windows. 

The temperature thresholds of $10^\circ$C and $30^\circ$C are standard agronomic indicators of cold and heat stress conditions affecting plant physiology and, in extreme cases, leading to growth inhibition or tissue damage~\cite{hatfield2015temperature}.
The same set of cumulative features is computed for both historical and future covariates.

Future meteorological covariates are treated as forecast-available inputs; in this retrospective study, observed daily meteorological variables serve as proxies for weather forecasts.
To account for forecast uncertainty, future covariates are perturbed during training with horizon-dependent multiplicative noise scaled by  $g_k = 1 + \beta \cdot \Delta t_k$, where $\Delta t_k$ is the temporal distance in days between the $k$-th future timestamp and the last historical observation. 
$\beta$ is set so that $g_K = 2$ at the final horizon step, meaning the perturbation intensity doubles over the forecasting window. 
The base perturbation is set to 10\% of the variable magnitude, a value selected via sensitivity analysis on the validation set.

All variables are standardized using training-set statistics and then transformed with $\operatorname{arcsinh}$ to reduce the influence of outliers, following~\cite{ansari2025chronos}.

\section{Methods}
The proposed transformer-based model, depicted in~\autoref{fig:methods}, performs quantile forecasting through two decoupled branches: a history encoder for past target values and covariates, and a future encoder for known future covariates.
Each input sequence is projected through a linear embedding layer into a shared latent space of dimension $d_{\text{model}}$, and positional encodings are added to preserve temporal ordering.
Missing NDVI targets, arising at overpasses fully obscured by clouds, are handled via binary masks that prevent the self-attention layers from attending to unobserved positions in both branches. 

The history encoder output is aggregated via \emph{Masked Temporal Average Pooling} to produce a compact representation of the past context.
For the future branch, the full covariate sequence is encoded via self-attention to allow all intermediate timesteps to inform the representations; only embeddings at actual Sentinel-2 acquisition times are then selected via \emph{Sparse Temporal Selection}, since forecasting targets exist only at those timestamps.

The pooled history representation is concatenated with the selected future embeddings along the feature dimension and passed to a quantile head that produces estimates at $q \in \{0.1, 0.5, 0.9\}$ for each future timestep.
The median ($q = 0.5$) serves as the point forecast; the $10^{th}$ and $90^{th}$ percentiles characterize predictive uncertainty. 
All future timesteps are predicted in parallel, without autoregressive decoding.

The model is trained with the quantile pinball loss, defined for 
quantile level $q_j \in (0,1)$ as:
\begin{equation}
\mathcal{L}_{q_j} =
|e|\Big(
q_j \cdot \mathbb{I}(e \ge 0)
+ (1 - q_j)\cdot \mathbb{I}(e < 0)
\Big)
\end{equation}
where $\mathbf{e} = \mathbf{y}_i - \hat{\mathbf{y}}_{i,q_j}$, $y_i$ is the ground-truth target, 
$\hat{y}_{i,q_i}$ is the predicted quantile at level $q_i$, and 
$\mathbb{I}(\cdot)$ is the indicator function.

To account for irregular temporal spacing and horizon-dependent 
uncertainty, each future target is down-weighted according to its 
temporal distance from the last historical observation $\tau_t$. 
The horizon-dependent weight is:
\begin{equation}
w_k = \frac{1}{1 + \alpha \cdot (\tau_{t+k} - \tau_t)}, 
\quad k = 1, \dots, h
\end{equation}
where the distance is expressed in days and $\alpha$ controls the 
decay rate. The weighted training loss is:
\begin{equation}
\mathcal{L} = \sum_{k=1}^{h} w_k \sum_{j=1}^{3} \mathcal{L}_{q_j}^{(k)}
\end{equation}
$\alpha$ is set to 0.1, which retains adequate supervision at longer 
horizons while emphasizing nearer, less uncertain targets.

\begin{table*}
\caption{Clear-sky NDVI forecasting results (mean $\pm$ std) for the proposed model and baselines. 
The proposed model is highlighted in \colorbox{lightblue}{blue}. \textbf{Bold} values denote the best results. Param and MFLOPs report model size and computational cost. Symbols indicate model families: \STAT\ Statistical, \REC\ Recurrent, \CONV\ Convolutional, \LLMcat\ LLM-based, \TRANS\ Transformer}
\centering
\small
%\resizebox{\columnwidth}{!}{
\begin{tabular}{lllllll|cc}
%\begin{tabular}{l llllll}
\toprule
\textbf{Architecture} & 
\bfseries RMSE $\downarrow$ &
\bfseries MAE $\downarrow$ &
\bfseries WMAPE $\downarrow$ &
\bfseries MASE $\downarrow$ &
\bfseries CRPS $\downarrow$ &
\bfseries Pinball $\downarrow$ &
\textbf{Param} &
\textbf{MFLOPs} \\
\midrule

\STAT~AutoARIMA
& $0.131_{\pm 0.274}$ &
$0.082_{\pm 0.103}$ &
$0.173_{\pm 0.217}$ &
$1.096_{\pm 1.371}$ &
$0.055_{\pm 0.071}$ &
$0.096_{\pm 0.045}$
& --
& --\\

\REC~LSTM
& $0.122_{\pm 0.150}$ 
& $0.088_{\pm 0.085}$ 
& $0.185_{\pm 0.180}$ 
& $1.171_{\pm 1.142}$ 
& $0.055_{\pm 0.060}$ 
& $0.094_{\pm 0.016}$
& 3.99 M
& 208.20
\\

\REC~RNN
 & $0.134_{\pm 0.146}$ 
& $0.099_{\pm 0.090}$ 
& $0.210_{\pm 0.191}$ 
& $1.327_{\pm 1.208}$ 
& $0.063_{\pm 0.064}$ 
& $0.109_{\pm 0.007}$
& 0.99 M
& 51.94
\\

 \REC~DeepAR
 & $0.116_{\pm 0.170}$ &
$0.075_{\pm 0.088}$ &
$0.158_{\pm 0.186}$ &
$1.000_{\pm 1.178}$ &
$0.050_{\pm 0.067}$ &
$0.088_{\pm 0.049}$ 
& 1.01 M
& 412.99
\\

\CONV~InceptionTime
& $0.123_{\pm 0.130}$ 
& $0.092_{\pm 0.081}$ 
& $0.195_{\pm 0.171}$ 
& $1.234_{\pm 1.083}$ 
& $0.068_{\pm 0.064}$ 
& $0.125_{\pm 0.168}$
& 1.23 M
& 64.42
\\
\LLMcat~TimeLLM
& $0.113_{\pm 0.160}$ &
$0.075_{\pm 0.085}$ &
$0.159_{\pm 0.179}$ &
$1.007_{\pm 1.130}$
& --
& --
& 124.44 M
& 510.61 
\\

\TRANS~PatchTST
& $0.127_{\pm 0.136}$ 
& $0.091_{\pm 0.088}$ 
& $0.191_{\pm 0.187}$ 
& $1.210_{\pm 1.181}$ 
& $0.054_{\pm 0.054}$ 
& $0.081_{\pm 0.015}$
& 3.18 M 
& 722.42
\\
\TRANS~Chronos-2
& \(0.141_{\pm0.233}\)
& \(0.086_{\pm0.111}\)
& \(0.181_{\pm0.235}\)
& \(1.147_{\pm1.488}\)
& \(0.057_{\pm0.069}\)
& \(0.033_{\pm0.040}\) 
& 28 M
& 316.87
\\
\TRANS 
\cellcolor{lightblue} \textbf{Ours}
& \cellcolor{lightblue}$\mathbf{0.096}_{\pm \mathbf{0.155}}$
& \cellcolor{lightblue}$\mathbf{0.062}_{\pm \mathbf{0.074}}$
& \cellcolor{lightblue}$\mathbf{0.130}_{\pm \mathbf{0.156}}$
& \cellcolor{lightblue}$\mathbf{0.828}_{\pm \mathbf{0.989}}$
& \cellcolor{lightblue}$\mathbf{0.038}_{\pm \mathbf{0.050}}$
& \cellcolor{lightblue}$\mathbf{0.021}_{\pm \mathbf{0.030}}$
& 2.16 M
& 111.97 \\
\bottomrule
\end{tabular}
%}
\label{tb:metrics_NDVI_clear_sky}
\end{table*}

\begin{table*}
\caption{Ablation study on temporal loss weighting ($w_k$) and meteorological feature engineering (FT-Eng) for NDVI forecasting (mean $\pm$ std). $\cmark$ indicates presence and $\xmark$ absence. \textbf{Bold} values denote the best results.}
\small
\centering
%\resizebox{\textwidth}{!}{
\begin{tabular}{cc|llllll}
\toprule
\bm{$w_k$} &
\textbf{ft-eng} &
\bfseries RMSE $\downarrow$ &
\bfseries MAE $\downarrow$ &
\bfseries WMAPE $\downarrow$ &
\bfseries MASE $\downarrow$ &
\bfseries CRPS $\downarrow$ &
\bfseries Pinball $\downarrow$ \\
\midrule

% --- NDVI clear sky ---
\xmark 
& \xmark
& \(0.099_{\pm0.158}\)
& \(0.064_{\pm0.076}\)
& \(0.134_{\pm0.161}\)
& \(0.854_{\pm1.022}\)
& \(0.040_{\pm0.053}\)
& \(0.022_{\pm0.032}\) \\

% --- Model unweighted but with ft-eng --
\xmark
& \cmark
& \(0.097_{\pm0.155}\)
& \(0.062_{\pm0.074}\)
& \(0.132_{\pm0.156}\)
& \(0.838_{\pm0.992}\)
& \(0.039_{\pm0.051}\)
& \(0.022_{\pm0.031}\)
\\

% --- Model weighted but w/ ft-eng --
\cmark 
& \xmark
& $0.098_{\pm 0.161}$ &
$0.063_{\pm 0.075}$ &
$0.132_{\pm 0.159}$ &
$0.842_{\pm 1.009}$ &
$0.038_{\pm 0.051}$ &
$0.022_{\pm 0.030}$ 
\\

% --- Model weighted and w ft-eng - alfa=0.1
\cmark
& \cmark
& $\mathbf{0.096}_{\pm \mathbf{0.155}}$ &
$\mathbf{0.062}_{\pm \mathbf{0.074}}$ &
$\mathbf{0.130}_{\pm \mathbf{0.156}}$ &
$\mathbf{0.828}_{\pm \mathbf{0.989}}$ &
$\mathbf{0.038}_{\pm \mathbf{0.050}}$ &
$\mathbf{0.021}_{\pm \mathbf{0.030}}$ \\

\bottomrule
\end{tabular}
%}
\label{tab:param_variation}
\end{table*}

\section{Experimental Configuration}
\paragraph*{\textbf{Evaluation metrics}}

We report standard point-forecast errors (RMSE, MAE, WMAPE, and MASE)~\cite{tortora2023matnet} computed on the median prediction ($q=0.5$).
For probabilistic assessment, we use the Continuous Ranked Probability Score (CRPS), which compares the predicted CDF $F(\cdot)$ with the observation $y$:
\begin{equation}
\mathrm{CRPS}(F,y) = \int_{-\infty}^{+\infty}\left(F(z)-\mathbb{I}(z \ge y)\right)^2\,dz.
\end{equation}
We also report the Pinball loss, averaged across quantile levels $q \in \{0.1, 0.5, 0.9\}$. 
Lower values indicate better performance for all metrics. 

\paragraph*{\textbf{Competitors}}
We compare the proposed model against baselines from five modeling paradigms: statistical (AutoARIMA~\cite{hyndman2008automatic}), recurrent (LSTMPlus~\cite{hochreiter1997long}, RNNPlus~\cite{elman1990finding}, DeepAR~\cite{salinas2020deepar}), convolutional (InceptionTimePlus~\cite{ismail2020inceptiontime}), LLM-based (TimeLLM~\cite{jin2023time}), and transformer-based (PatchTST~\cite{nie2022time}, Chronos-2~\cite{ansari2025chronos}). 
LSTMPlus, RNNPlus, InceptionTimePlus, and PatchTST are implemented via the tsai library~\cite{tsai}; DeepAR, AutoARIMA, and Chronos-2 via AutoGluon~\cite{agtimeseries}; TimeLLM via NeuralForecast~\cite{olivares2022library_neuralforecast}. 
Chronos-2 is evaluated zero-shot to assess its out-of-the-box generalization to the NDVI forecasting task.
All baselines operate in the multivariate setting, except TimeLLM and AutoARIMA, which use target history only.

\paragraph*{\textbf{Implementation details}}
The model is implemented in PyTorch and trained with the Adam optimizer for 200 epochs (batch size 128, initial learning rate $10^{-4}$).
A validation set corresponding to 20\% of the training data is used for model selection.
The learning rate is reduced by a factor of 0.2 after 20 epochs without improvement on the validation loss, down to a minimum of $5\times10^{-5}$.
Both the history and future branches use identical transformer encoders with 8 layers, $d_{\text{model}}=128$, 8 attention heads, and a 512-dimensional feed-forward network.
Dropout ($p=0.1$) is applied throughout the transformer blocks.
The training set contains 55{,}341 samples and the test set 3{,}336 samples.
All experiments are conducted on an NVIDIA T4 GPU.

\begin{table*}[ht]
\caption{Ablation study on forecasting models. Each configuration enables (\cmark) or disables (\xmark) conditioning on historical meteorological covariates, future meteorological covariates, and past target values. \textbf{Bold} values denote the best results.}
\centering
\small
\resizebox{0.85\textwidth}{!}{
\begin{tabular}{cccccccccc}
\toprule
\bfseries Future &
\bfseries History &
\bfseries Target &
\bfseries RMSE $\downarrow$ &
\bfseries MAE $\downarrow$ &
\bfseries WMAPE $\downarrow$ &
\bfseries MASE $\downarrow$ &
\bfseries CRPS $\downarrow$ &
\bfseries Pinball $\downarrow$ \\
    \midrule

% Future-Cov
\xmark & \cmark & \cmark
& $0.104_{\pm 0.172}$ &
$0.067_{\pm 0.079}$ &
$0.142_{\pm 0.167}$ &
$0.905_{\pm 1.062}$ &
$0.042_{\pm 0.056}$ &
$0.024_{\pm 0.034}$ \\

% Target-Hist
\cmark & \cmark & \xmark
& $0.154_{\pm 0.133}$ &
$0.118_{\pm 0.099}$ &
$0.249_{\pm 0.208}$ &
$1.581_{\pm 1.324}$ &
$0.072_{\pm 0.066}$ &
$0.040_{\pm 0.039}$ \\

% Hist-Cov
\cmark & \xmark & \cmark
& $0.099_{\pm 0.157}$ &
$0.064_{\pm 0.076}$ &
$0.136_{\pm 0.160}$ &
$0.864_{\pm 1.016}$ &
$0.040_{\pm 0.052}$ &
$0.022_{\pm 0.031}$ \\

% Future + History
\xmark & \xmark & \cmark
& $0.104_{\pm 0.162}$ &
$0.067_{\pm 0.079}$ &
$0.141_{\pm 0.167}$ &
$0.897_{\pm 1.064}$ &
$0.040_{\pm 0.051}$ &
$0.023_{\pm 0.030}$ \\

% Future + Target (FutureCov + Targetd)
\xmark & \cmark & \xmark
& $0.165_{\pm 0.135}$ &
$0.129_{\pm 0.103}$ &
$0.273_{\pm 0.218}$ &
$1.734_{\pm 1.385}$ &
$0.078_{\pm 0.069}$ &
$0.044_{\pm 0.041}$ \\

% Hist + Target
\cmark & \xmark & \xmark
& $0.161_{\pm 0.138}$ &
$0.124_{\pm 0.103}$ &
$0.262_{\pm 0.218}$ &
$1.664_{\pm 1.385}$ &
$0.076_{\pm 0.070}$ &
$0.042_{\pm 0.041}$ \\

% Ours with alfa= 0.1
\cmark & \cmark & \cmark
& $\mathbf{0.096}_{\pm \mathbf{0.155}}$ &
$\mathbf{0.062}_{\pm \mathbf{0.074}}$ &
$\mathbf{0.130}_{\pm \mathbf{0.156}}$ &
$\mathbf{0.828}_{\pm \mathbf{0.989}}$ &
$\mathbf{0.038}_{\pm \mathbf{0.050}}$ &
$\mathbf{0.021}_{\pm \mathbf{0.030}}$ \\

\bottomrule
\end{tabular}
}
\label{tab:ablation_study}
\end{table*}

\begin{figure}
    \small
    \centering
    \includegraphics[width=\columnwidth]{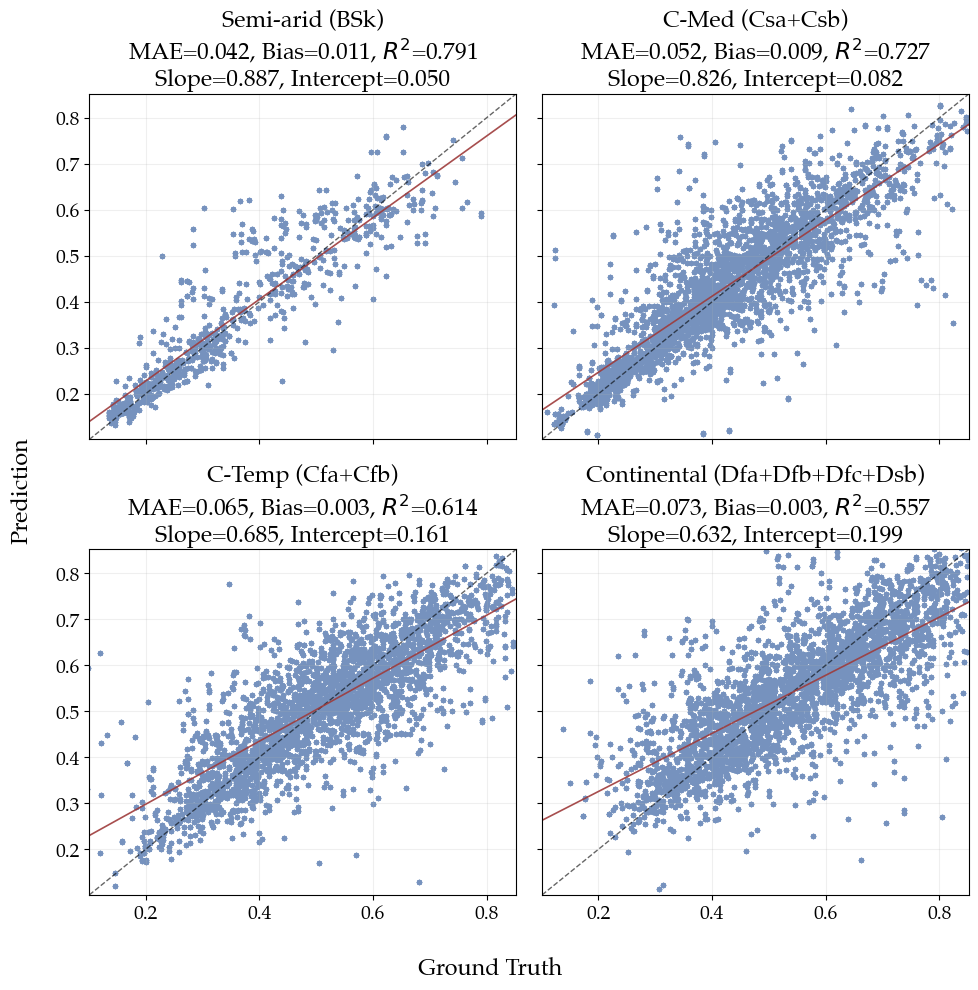}
    \caption{Ground truth versus predicted NDVI by aggregated K{\"o}ppen--Geiger climate group. 
    The black dashed line shows the 1:1 reference, and the red line shows the linear fit.}
    \label{fig:gt_vs_pred}
\end{figure}

\begin{figure}
    \small
    \centering
    \includegraphics[width=\columnwidth]{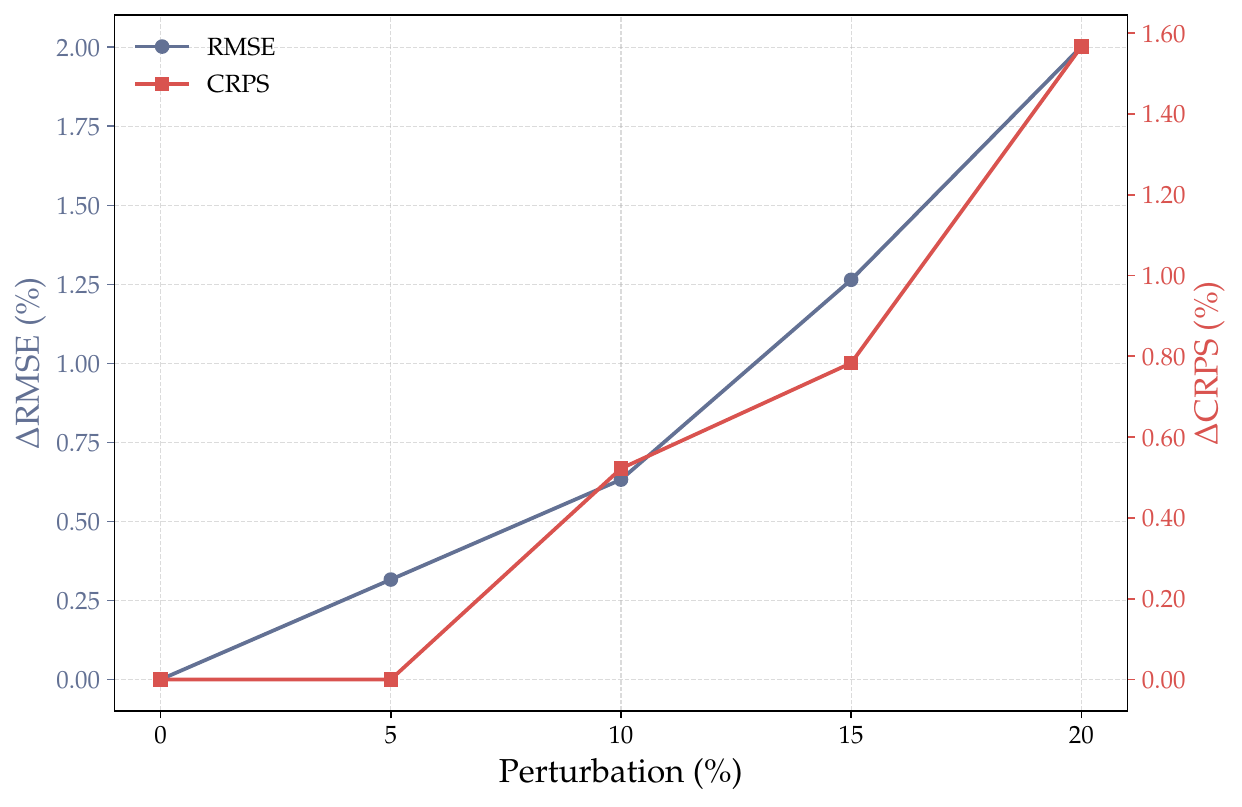}
    \caption{Performance variation under increasing noise in future covariates. Results are reported as percentage variation with respect to the no-noise baseline (0\% perturbation) for RMSE (left axis) and CRPS (right axis).}
    \label{fig:noise_robustness}
\end{figure}

\section{Results}
We evaluate the proposed model against all baselines on the clear-sky NDVI forecasting task, followed by two ablation studies, a stratified analysis across K{\"o}ppen--Geiger climate groups, and a noise sensitivity analysis.

\paragraph*{\textbf{Comparative analysis}}
~\autoref{tb:metrics_NDVI_clear_sky} reports forecasting performance on the clear-sky NDVI task for all models under the same experimental protocol.
TimeLLM produces deterministic forecasts; therefore, CRPS and Pinball loss are not reported.
The proposed model requires only 2.16~M parameters and 111.97~MFLOPs, remaining substantially lighter than models such as TimeLLM and Chronos-2. 
It outperforms smaller recurrent baselines such as LSTM, RNN, and DeepAR, indicating that the performance gain is not attributable to model capacity alone. 
The proposed model achieves the best results across all metrics.
Statistical significance is assessed using the Diebold--Mariano test with a Newey--West heteroskedasticity and autocorrelation consistent variance correction~\cite{diebold2002comparing}. 
The truncation lag is set to the three-step forecasting horizon. 
All pairwise comparisons between the proposed model and each baseline yield $< 0.001$, confirming that the observed improvements are statistically significant.

\paragraph*{\textbf{Ablation study on temporal loss weighting and feature engineering}}
\autoref{tab:param_variation} evaluates the impact of temporal loss weighting ($w_k$) and engineered meteorological features (\textit{ft-eng}). 
Disabling $w_k$ corresponds to uniform weighting of future errors in the Pinball loss. 
Both components improve pointwise and probabilistic metrics. 
Temporal weighting provides the largest gains, in line with its role in handling irregular temporal spacing and horizon-dependent uncertainty. 
Feature engineering yields smaller but systematic improvements, indicating that cumulative and extreme-weather descriptors complement raw meteorological covariates.

\paragraph*{\textbf{Ablation study on input modalities}} 
\autoref{tab:ablation_study} assesses the contribution of target history, historical meteorological covariates, and future meteorological covariates.
Removing target history produces the largest degradation across pointwise and probabilistic metrics, confirming that recent vegetation dynamics provide the dominant signal at short horizons.
Historical meteorological covariates provide secondary gains when combined with target history.
Future covariates alone are insufficient to achieve competitive performance, but contribute incrementally in the full multimodal setting.
%The results indicate a clear hierarchy among modalities: target history is the primary signal, historical covariates provide secondary gains, and future covariates contribute incrementally in the full multimodal setting.

\paragraph*{\textbf{NDVI prediction across climate regions}}
\autoref{fig:gt_vs_pred} reports ground truth versus predicted NDVI stratified by aggregated K{\"o}ppen--Geiger climate groups.
This aggregation was introduced to reduce class fragmentation and better reflect the European study area.
Across all groups, predictions align closely with the 1:1 reference line, though performance degrades from semi-arid to continental regimes: MAE increases from 0.042 (Semi-arid, BSk) to 0.073 (Continental, Dfa+Dfb+Dfc+Dsb), while $R^2$ (fraction of explained variance) decreases from 0.791 to 0.557. 
The mean bias is small and positive across all groups (0.003--0.011), indicating a secondary overestimation tendency relative to the dominant error component.
C-Med, C-Temp and Continental climates show a broader spread at intermediate NDVI levels, attributable to higher phenological variability and observation noise from persistent cloud cover and seasonal effects.

\paragraph*{\textbf{Noise sensitivity analysis}} \autoref{fig:noise_robustness} reports the effect of increasing perturbations on future meteorological covariates, ranging from 0\%  to 20\% in 5\% increments, simulating imperfect weather forecasts.  Both RMSE and CRPS are expressed as percentage change relative to the no-noise baseline. Degradation is moderate up to 10\%, beyond which both metrics increase more sharply, with RMSE reaching $+2\%$ and  CRPS $+1.6\%$ at 20\% perturbation. This behavior supports the choice of 10\% as the base perturbation level during training.

\section{Conclusion} 
This work presents a probabilistic framework for short-term NDVI forecasting from sparse and irregular clear-sky satellite time series, integrating historical vegetation dynamics with meteorological covariates through a transformer-based architecture.
Experiments on large-scale European data show that the proposed model outperforms statistical, deep learning, and time-series baselines on both pointwise and probabilistic metrics.
Ablation studies confirm that target history is the primary driver of performance, while meteorological covariates provide additional gains in the full multimodal setting. 
Stratified analyses across K{\"o}ppen--Geiger climate groups show that performance degrades gracefully from semi-arid to continental regimes.
The evaluation is restricted to European agricultural areas, and generalization to other regions remains to be assessed, as crop phenology, management practices, and climate regimes may differ substantially. 
Future work will extend the framework by incorporating explicit climate-zone and crop-type conditioning to improve generalization and robustness across heterogeneous agricultural landscapes.

\section{Acknowledgments}
Irene Iele and Daniele Molino are Ph.D. students enrolled in the National Ph.D. in Artificial Intelligence, Health and Life Sciences, organized by Università Campus Bio-Medico di Roma.
We acknowledge i) the EuroHPC Joint Undertaking for granting this project access to the EuroHPC supercomputer Vega, hosted by the Institute of Information Science (Slovenia), under a EuroHPC Development Access call ii) Project ECS 0000024 Rome Technopole, - CUP C83C22000510001,  NRP Mission 4 Component 2 Investment 1.5,  Funded by the European Union - NextGenerationEU.

%\IEEEtriggeratref{22}
\bibliographystyle{IEEEtran}
\bibliography{references}
\end{document}